\documentclass[10pt,twocolumn,letterpaper]{article}

\usepackage{cvpr}   % To produce the CAMERA-READY version
\usepackage{adjustbox}
\usepackage{xcolor}
\usepackage{comment}
\definecolor{cvprblue}{rgb}{0.21,0.49,0.74}
\usepackage[pagebackref,breaklinks,colorlinks,allcolors=cvprblue]{hyperref}
\usepackage{ulem} % For strikethrough
\usepackage{amssymb}
\usepackage{algorithm}
\usepackage{algpseudocode}
\usepackage{caption, threeparttable}
\usepackage{booktabs}
\usepackage{enumitem}

\def\paperID{6} % *** Enter the Paper ID here
\def\confName{CVPR}
\def\confYear{2025}

\title{Perceiving Beyond Language Priors: Enhancing Visual Comprehension and Attention in Multimodal Models}

\author{Aarti Ghatkesar\\
{\tt\small aarti@cerebras.net}
\and
\\
\\
\textsc{AppliedML, Cerebras}
\and
Ganesh Venkatesh\\
{\tt\small ganesh.venkatesh@cerebras.net}
}

\begin{document}
\maketitle
\newcommand{\ignore}[1]{}
\newcommand{\fix}[1]{#1}
\definecolor{DarkGreen}{RGB}{1,50,32}
\newcommand{\hl}[1]{\textbf{\textcolor{DarkGreen}{#1}}}
\newcommand{\ud}[1]{\textcolor{purple}{#1}}
\newcommand{\vl}{\scalebox{.9}[1.0]{\textsc{VisualLoss}}}
\newcommand{\vla}{\scalebox{.9}[1.0]{\textsc{VisualLossAdv}}}
\newcommand{\bt}{\scalebox{.9}[1.0]{\textsc{BlankTokens}}}
\newcommand{\our}{\scalebox{.9}[1.0]{\textsc{PerceptLLM}}}
\newcommand{\ours}{\scalebox{.9}[1.0]{\textsc{PerceptLLM+Syn}}}
\newcommand{\algcomment}[1]{\scalebox{.8}[1.0]{\textit{\texttt{#1}}}}

\begin{abstract}
Achieving deep alignment between vision and language remains a central challenge for Multimodal Large Language Models (MLLMs). These models often fail to fully leverage visual input, defaulting to strong language priors. Our approach first provides insights into how MLLMs internally build visual understanding of image regions and then introduces techniques to amplify this capability. We explore techniques designed both to deepen the model's understanding of visual content (\vl) and to ensure that these visual insights actively guide language generation (\bt). We demonstrate the superior multimodal understanding of our resultant model, \our, through a detailed upstream analysis on predicting visually-dependent tokens as well as $>$\fix{10 percentage point} boost on a visually challenging task and a consistent boost across multiple tasks.\footnote{Accepted to the Second Workshop on Visual Concepts at CVPR 2025.}
\end{abstract}    
\vspace{-15pt}
\section{Introduction}
\label{sec:intro}

\ignore{Outline: i) Multimodal models have demonstrated very impressive general-purpose visual QA capabilities, ii) Recent analysis have shown that multimodal models heavily rely on language priors and can be very weak at leveraging fine-grained visual signals, iii) We propose techniques to address these challenges using a three-pronged approach that strengthens visual signals, encourages model to pay more attention to visual input when generating response and construct synthetic dataset that leverages our advanced training fabric}

\ignore{First para - background}
Recent years have witnessed remarkable progress in the field of multimodal artificial intelligence, particularly with the advent of Large Multimodal Models (MLLMs)~\cite{llava,mllama,pixtral,mllmsurvey}. These models, capable of jointly processing information from different modalities like vision and language, have demonstrated impressive general-purpose capabilities across a wide range of tasks. \ignore{Their proficiency in areas such as visual question answering~\cite{vqabench}, image captioning~\cite{llavabench}, and multimodal dialogue~\cite{diagbench} is particularly noteworthy, often generating remarkably coherent and contextually relevant responses that suggest a deep understanding of visual scenes paired with sophisticated language generation abilities.} This success has spurred significant interest and research, pushing the boundaries of what AI systems can perceive, understand, and communicate about the visual world.

\ignore{Second para -- challenge}

Despite these striking capabilities, recent work on probing the workings of MLLMs have begun to uncover significant limitations in how current MLLMs integrate visual information~\cite{spatialeval,mllmhallucinate,vllmblind}. This growing body of evidence reveals that many state-of-the-art models exhibit a strong reliance on learned language priors and statistical correlations present in their vast training datasets. Consequently, they can be surprisingly weak at leveraging fine-grained visual signals crucial for nuanced understanding. Models may generate plausible-sounding but factually incorrect responses when faced with visual details that contradict common knowledge or require careful observation, effectively `hallucinating' or ignoring specific visual evidence in favor of more probable textual outputs. This gap highlights a critical challenge: ensuring that visual input genuinely grounds the model's reasoning and generation processes. 

%%\ignore{third para -- our proposal}
To address these fundamental challenges, this paper proposes \our, multimodal models trained with a novel methodology centered on enhancing the model's internal visual representation. 
First, we strengthen the visual signal processing,
enabling the model to extract richer and more fine-grained visual representations. Second, we introduce a training mechanism that explicitly encourages the model to allocate greater attention to these enhanced visual inputs during the response generation phase, reducing over-reliance on language priors. Finally, complementing these training modifications, we construct a targeted synthetic dataset specifically designed to leverage this advanced training fabric, providing controlled examples that force the model to learn and exploit fine-grained visual cues effectively. 
%%
%%\ignore{Contributions}
%%
Our key contributions are:
% \vspace{-2pt}

% \begin{itemize}[nosep,leftmargin=*]
\noindent \textbf{Enhanced Visual Representation via Auxiliary Loss and Disentangled Architecture.} We introduce a novel approach that combines an auxiliary visual loss (\vl) with a key architectural modification. The loss function ensures the LLM backbone builds a rich representation of the entire input image, independent of associated text, while our architectural change decouples the processing pathways for image and text tokens using independent weights. This disentanglement allows the model to specialize in rich visual representation learning without conflicting with its autoregressive text generation objective.

\noindent \textbf{Weakening Language-prior via \bt.} To fully leverage the model's enhanced visual understanding, we introduce a technique to gently reduce the LLM backbone's reliance on strong language priors. Complementing this, we develop a specialized synthetic dataset, specifically to encourage sensitivity to fine-grained visual details.

\noindent \textbf{Analysis of Multimodal Alignment.} We provide insights into how MLLMs attempt to understand visual information and motivates our approach. Our analysis demonstrates significant improvements in multimodal alignment, showcasing that our work pushes MLLMs towards a more visually faithful reasoning and generation.
% \end{itemize}

% \begin{itemize}
%     \item \textbf{Enhanced Visual Representation using \vl:} We introduce a new loss function to ensure the LLM backbone builds a rich internal representation of all aspects of the input image, independent of whether they are mentioned in associated training text.
%     \item \textbf{Reducing Reliance on Language Priors using \bt:} To fully leverage the model's enhanced visual understanding, we introduce a technique to gently reduce the LLM backbone's reliance on strong language priors. Complementing this, we develop a specialized synthetic dataset, constructed specifically to encourage heightened sensitivity to fine-grained visual details.
%     \item \textbf{Analysis of Multimodal Alignment:} We provide insights into how MLLMs attempt to understand visual information and motivates our approach. Our analysis demonstrates significant improvements in multimodal alignment, showcasing that our work pushes MLLMs towards a more visually faithful reasoning and generation.
% \end{itemize}

\ignore{
\begin{figure*}
  \centering
  \begin{subfigure}{0.68\linewidth}
    \fbox{\rule{0pt}{2in} \rule{.9\linewidth}{0pt}}
    \caption{An example of a subfigure.}
    \label{fig:short-a}
  \end{subfigure}
  \hfill
  \begin{subfigure}{0.28\linewidth}
    \fbox{\rule{0pt}{2in} \rule{.9\linewidth}{0pt}}
    \caption{Another example of a subfigure.}
    \label{fig:short-b}
  \end{subfigure}
  \caption{Example of a short caption, which should be centered.}
  \label{fig:short}
\end{figure*}
}
\section{Challenges and Related Work}
\label{sec:cr}

This section first overviews the key challenges in multimodal LLMs that we seek to address in our work and then discuss prior work on addressing these challenges.

\subsection{Challenges in Multimodal LLMs}
\label{sec:challenge}

\ignore{Two challenges: i) Prior work such as Eyes wide shut, Hallucinations of MLLM survey among others identify weak visual understanding because CLIP does not have a detailed understanding of the image, ii) Prior work such as SpatialEvals demonstrate LLM does not know how to leverage visual data to answer question. }

We highlight three key challenges from prior works and address them in this paper:

\noindent \textbf{Weak Visual Understanding.} First, several studies highlight the often weak visual understanding capabilities inherent in many contemporary MLLMs~\cite{eyeswideshut, mllmhallucinate}. This limitation is frequently attributed, at least in part, to the nature of the vision encoders employed, such as CLIP. While effective for capturing global image-text semantics, encoders like CLIP may not provide the sufficiently detailed, fine-grained representations required for nuanced visual comprehension, leading to models struggling with specific object attributes, states, or intricate scene details.

\noindent \textbf{Sparse Training Signals.} Second fundamental challenge limiting the depth of visual understanding in current MLLM arises from the sparse nature of the loss signals. The standard next-token prediction loss is calculated exclusively on the text sequence which provides only an indirect and often weak signal for visual learning, as only a fraction of the text tokens may have a strong, unambiguous dependence on the visual content. Furthermore, the textual descriptions frequently refer only to a subset of the objects, attributes, and relationships present in the image. As a result of this sparse supervision derived from text, the LLM backbone has limited opportunity to develop a truly rich, comprehensive internal representation of the full visual context.

\noindent \textbf{Strong Language Priors.} As demonstrated in prior work like SpatialEval~\cite{spatialeval}, models may fail to utilize visual context correctly, particularly for tasks demanding spatial reasoning or precise grounding of textual concepts in the image. This indicates that even if relevant visual information is encoded, the mechanisms for cross-modal interaction and reasoning within the LLM component may be insufficient to properly access, interpret, and apply that information when answering questions or generating descriptions.

\subsection{Related Work}
\label{sec:related}

Prior work attempts to address the above challenges in multiple different ways. These include strengthening visual signals by providing embeddings from multiple visual encoders to the LLM backbone~\cite{cambrian,eyeswideshut}, adding auxiliary losses to help the LLM build a richer visual representation~\cite{llavaground}, and generating synthetic or augmented data~\cite{sparkle,grand}. Our training innovations fundamentally differ from these approaches in that we do not require extra visual annotations or the overhead of additional encoders at inference time (details in Appendix~\ref{sec:relatedworkappendix}).

\section{Our Approach: \our}
\label{sec:approach}

\ignore{\ud{Our multi-modal LLM integrates visual and textual inputs through a novel dual-encoder framework, optimized for joint representation learning. Below, we formalize the dataset structure, model components, and training protocol. Let $\mathcal{D} =\mathcal{I} \times \mathcal{T}$ represent the paired image ($\mathcal{I}$) and textual dataset ($\mathcal{T}$). $\mathcal{F}(\cdot; \theta_F)$ represent the trainable LLM backbone, and $\mathcal{G}(\cdot; \theta_G)$ represent the vision encoder that feeds into the LLM via a projector represented by $\mathcal{M}(\cdot; \theta_M)$.
}}

\ignore{Overview blurb -- i) Start with probing behavior of existing MLLMs to motivate our approach, ii) first training innovation \texttt{visual loss} that deepens LLM backbone's understanding of visual concepts at a per-patch level, iii) second innovation \texttt{blank tokens} that gently weakens LLM backbone's reliance on language priors and makes it rely more on the rich visual rep to produce response.}

We first analyze the internal visual understanding in existing MLLMs (Section~\ref{sec:motivation}) and rest of this section details our proposed training strategy shown in Algorithm~\ref{alg:approach}. 

% \subsection{\ud{\sout{Motivation:}} Dissecting visual tokens in MLLMs}
\subsection{Dissecting visual tokens in MLLMs}
\label{sec:motivation}
\begin{figure}
\centering
\includegraphics[width=0.95\columnwidth]{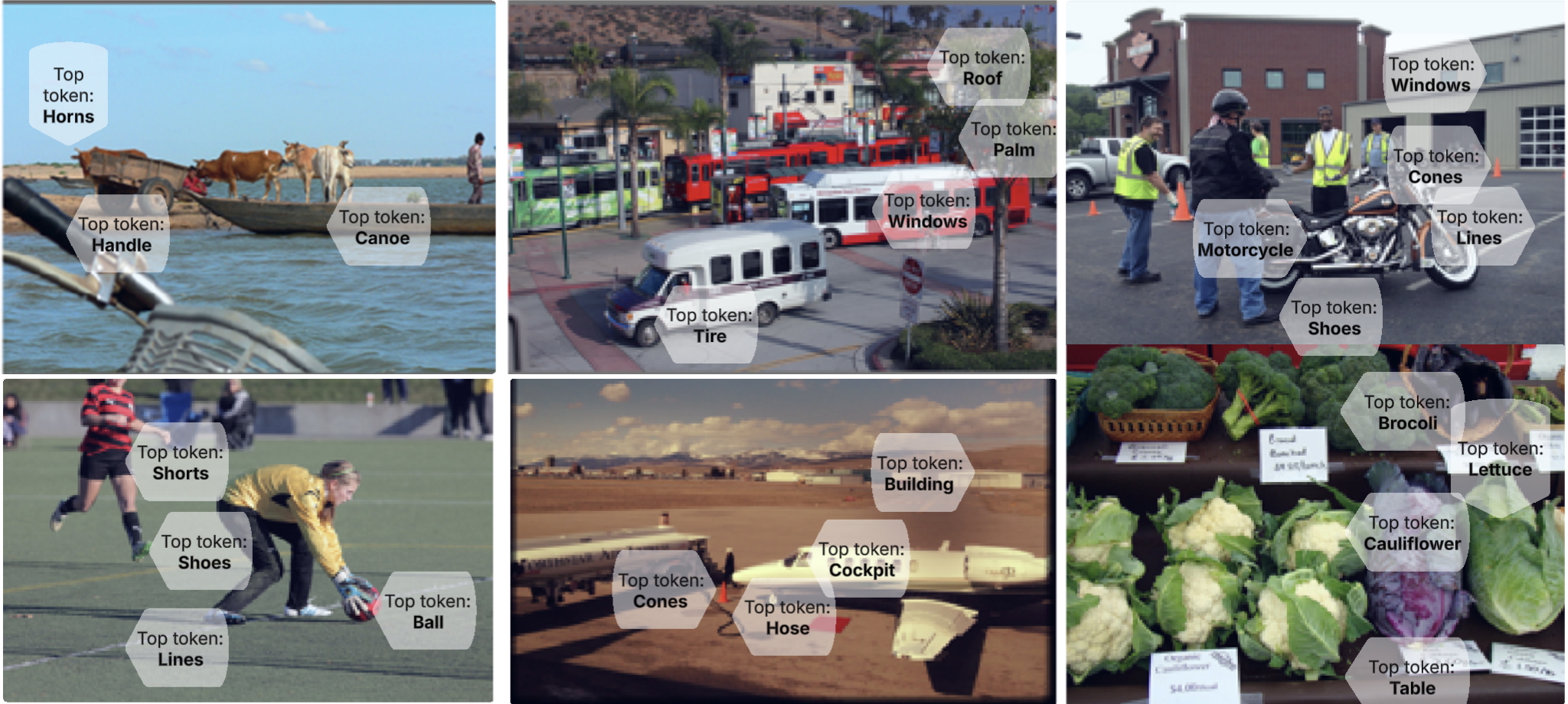}
\caption{\label{fig:motivation} Sample demonstrates how MLLMs attempt to build an internal representation that capture semantic information about the image patches even without any explicit supervision}
\end{figure}

Understanding how MLLMs internally develop visual representations is crucial for addressing their limitations and guiding future improvements. To gain insights into this process, we perform probing analyses focused specifically on the visual tokens processed by the model following initial encoding (Figure~\ref{fig:motivation}). In particular, we feed the visual token embeddings generated by the LLM backbone through its final layer, i.e., language modeling head (\texttt{lm\_head}) and identify the top predicted token for each visual patch. We qualitatively examined a diverse set of images and present examples in Figure~\ref{fig:motivation}. Some key observations include:
\begin{itemize}
    \item The model captures information about both foreground and background elements.
    \item The model is capable of localizing fine-grained structures or small items (e.g., \texttt{lines}, \texttt{cones}).
    \item The model captures \textit{semantically distinct} items at varying scales (e.g., \texttt{handle, canoe, cow}) within an image.
\end{itemize}

Our investigations indicate that MLLMs do not treat visual tokens merely as abstract feature vectors. Instead, they exhibit an emergent capability to associate meaningful semantic labels or concepts with their corresponding image patches. This suggests that the model performs a degree of localized scene interpretation prior to extensive cross-modal fusion. Notably, MLLMs develop this localized understanding without explicit supervision for this capability. 

This observation -- that MLLMs develop rudimentary semantic grounding at the visual token level -- provides strong motivation for our proposed approach. Given that the model possesses this nascent ability, interventions designed both to enhance this specific capability and to encourage the model to more effectively utilize this internal visual information hold significant potential. Our approach to achieving this goal is detailed in the following three sub-sections.

\subsection{Visual Representation using \vl}
\label{sec:visualloss}
\begin{figure}
\centering
\includegraphics[width=\columnwidth]{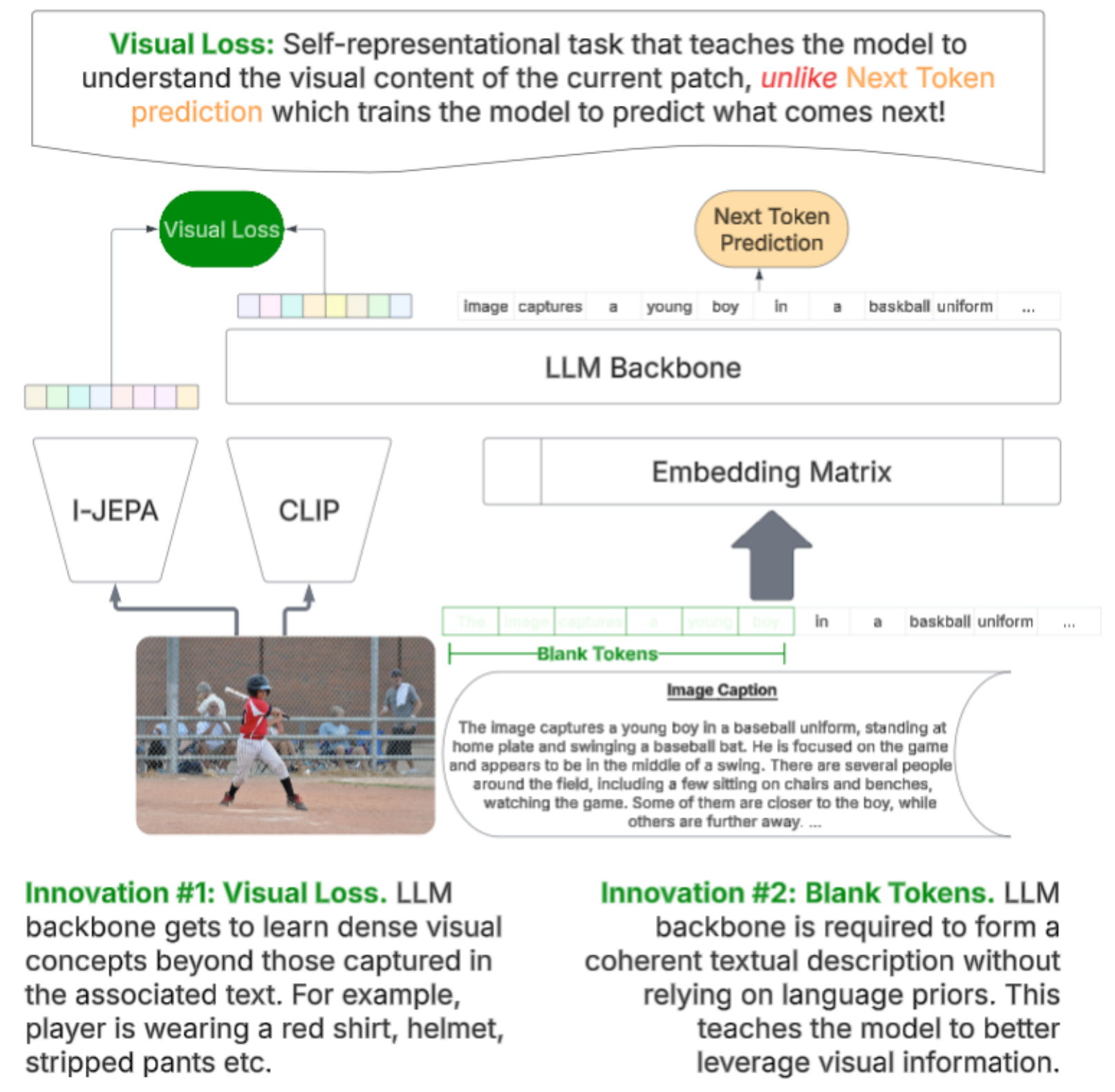}
\caption{\label{fig:approach} Proposed modifications to MLLM training that enhances visual understanding through \vl\ (Section~\ref{sec:visualloss}) and encourages LLM backbone to pay greater attention to them using \bt\ (Section~\ref{sec:blanktokens}). We also highlight the tension between \vl\ and next-token prediction that is addressed through independent pathways for vision and text (Section~\ref{sec:disentangled_weights}).}
\vspace{-5pt}
\end{figure}

\ignore{Previous techniques aimed at improving the visual understanding of MLLMs have explored methods such as feeding representations from multiple vision encoders as input to the core language model~\cite{whatsup,multiencoder2,multiencoder3}. Although seemingly promising, a key limitation stems from the nature of the standard training objective based on next-token prediction that primarily rewards textual fluency and prediction accuracy, providing only an indirect and weak incentive for the LLM to develop an understanding of the visual input's nuances and semantic content. An additional challenge is that visual encoders' latent spaces are often not aligned with the LLM's embedding space, making it even more difficult for LLMs to internalize these visual concepts. As a result, the LLM is often unable to fully leverage the visual richness provided by the additional visual encoders.

To overcome this, }

We introduce a fundamentally new strategy (referred to as \vl\ in this work) to enrich the LLM's internal processing rather than just augmenting its inputs, as illustrated in Figure~\ref{fig:approach}. In particular, we introduce an auxiliary vision encoder with rich visual semantic understanding  (I-JEPA~\cite{ijepa} in our work) during training phase and use it to introduce a new loss term, \vl, on the visual tokens of the LLM backbone. By requiring the LLM backbone to predict corresponding I-JEPA representations, we provide a strong, targeted supervisory signal that directly fosters a deeper and more nuanced visual understanding within the language model itself. This process encourages the model to build a comprehensive understanding of the visual content, rather than being limited to learning only those concepts explicitly captured in the text description. Please refer to Appendix~\ref{sec:visuallossappendix} for details.

\subsection{Language Priors using \bt}
\label{sec:blanktokens}

Achieving accurate and visually grounded responses requires addressing another key challenge: over-reliance of MLLMs on strong language priors. Even with improved visual feature processing, models often default to generating text driven by learned linguistic patterns, potentially overlooking specific visual evidence. To mitigate this tendency, we introduce a complementary strategy focused on gently recalibrating the model's dependence on textual context during training. This involves strategically masking portions of the input text, thereby disrupting straightforward language-based auto-completion and compelling the model to rely more heavily on its visual comprehension capabilities to generate coherent outputs. Please refer to Appendix~\ref{sec:blanktokensappendix} for details.
\begin{algorithm}
\caption{\label{alg:approach} Forward pass with proposed enhancements.}
\begin{algorithmic}[1]
\State \textbf{INPUT:} I, T$_{in}$, T$_{out}$, \texttt{MLLM}, V$_{aux}$, $\beta$
\State \textbf{OUTPUT:} \textsc{Loss}
\State \algcomment{\#I: Image, T$_{in}$, T$_{out}$: Text input and output}
\State \algcomment{\#V$_{aux}$: Aux Vision Encoder}
\State T$_{inBlank}$ = BlankInputsPartial(T$_{in}$)  \algcomment{\#Section~\ref{sec:blanktokens}}
\State V$_{feat}$, T$_{feat}$ = \texttt{MLLM}(I, T$_{in}$)
\State \algcomment{\#V$_{feat}$, T$_{feat}$ is visual and text features}
\State V$_{emb}$ = V$_{aux}$(I) \algcomment{\#Section~\ref{sec:visualloss}}
\State $\mathcal{L}$$_{ntp}$ = $\mathcal{CE}$(\texttt{lm\_head}(T$_{feat}$), T$_{out}$) \algcomment{\#Next token loss} 
\State $\mathcal{L}_{visualLoss}$ = VisualLoss(V$_{feat}$, V$_{emb}$) \algcomment{\#Section~\ref{sec:visualloss}}
\State $\mathcal{L}_{tot}$ = $\mathcal{L}_{ntp}$ + $\beta \cdot $ $\mathcal{L}_{visualLoss}$
\end{algorithmic}
\end{algorithm}

\subsection{Disentangling Visual Representation and Autoregressive Prediction}
\label{sec:disentangled_weights}

While the auxiliary \vl\ described previously proved effective in enhancing model's visual representation (Section~\ref{sec:results}), we also observe that it introduces an inherent tension in the LLM backbone's training objectives. The model is simultaneously tasked with two fundamentally different goals depending on the input modality. For text tokens, the objective is standard causal language modeling: each token's representation is optimized to predict the \textit{next} token in the sequence. This is an explicitly autoregressive task that relies on a natural left-to-right flow. In contrast, for image tokens, our visual loss encourages the model to learn a rich representation of the \textit{current} patch, as images lack the same inherent sequential or autoregressive structure. Forcing a single set of Transformer weights to handle both these conflicting, modality-dependent objectives leads to the weights being pulled in different directions.

To resolve this conflict, we introduce a key architectural modification: we decouple the processing of image and text tokens within the LLM backbone by assigning them independent sets of weights~\cite{mot,cogvlm}. By creating this programmatic distinction, we effectively allow for two specialized pathways to emerge within the Transformer layers. The pathway for image tokens can optimize its parameters, specifically the self-attention parameters, for the non-autoregressive task of building rich spatial representations. Concurrently, the pathway for text tokens can retain its well-established autoregressive capabilities for fluent and coherent language generation. This disentanglement allows for distinct attention behaviors tailored to each modality, enabling a more effective and specialized learning process for both visual understanding and textual prediction. 

As demonstrated empirically in our results (Section~\ref{sec:spatialevals}), this architectural modification yields a significant improvement in model accuracy, providing strong empirical validation for our hypothesis. Importantly, this approach maintains deployment efficiency. Despite the increase in the total number of parameters to support both modalities, the model incurs no additional computational or memory access overhead during inference, since only the pathway corresponding to the token's modality (image or text) is utilized at any given step.

\subsection{Targeted Synthetic Data Generation}
\label{sec:synthetic} % Label referenced in previous examples
\vspace{-5pt}

\ignore{Inspired by prior work on building synthetic data for building block capabilities, we construct a dataset that randomly samples objects from Open Web Images, places them on a NxN grid and asks basic questions about relative object direction and distance. These questions can only be answered by understanding the visual content and reasoning over them and so compliment our training innovations by providing the model with a target task to sharpen its modality alignment.}

\begin{figure}[!h]
\centering
\includegraphics[width=0.95\columnwidth]{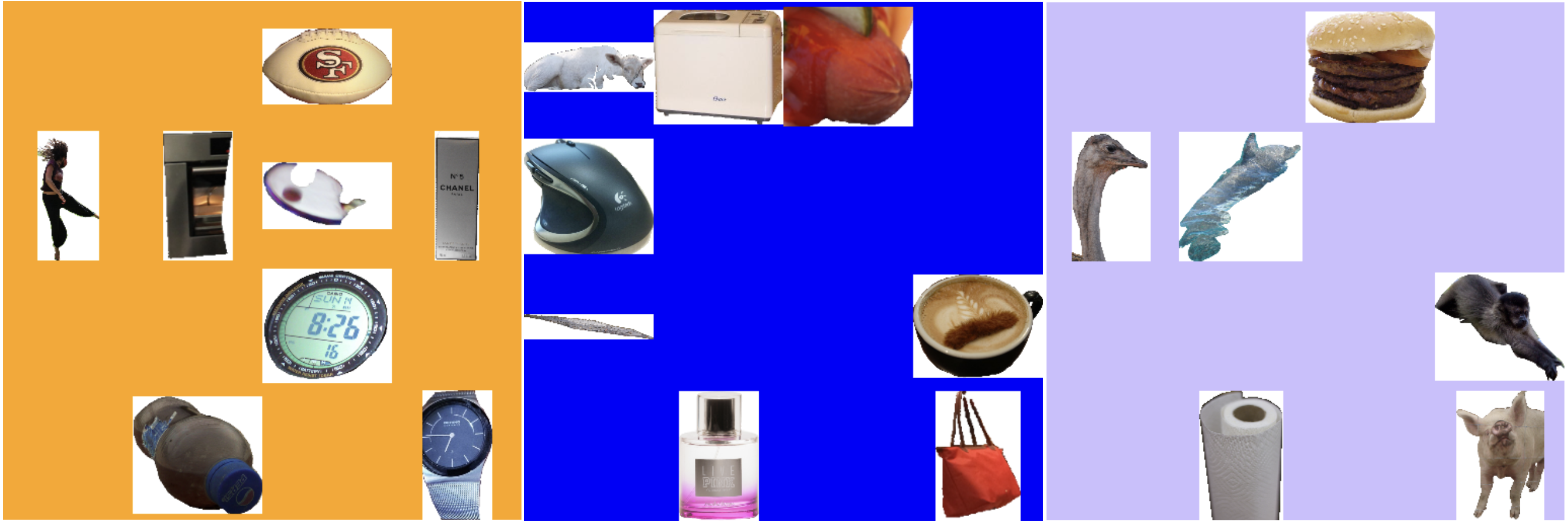}
\caption{\label{fig:syndata} Examples of synthetic grid data for spatial awareness. Objects are randomly sampled from a large public image collection~\cite{benenson2022colouring,kuznetsova2020open,openimages}, and placed onto distinct locations.}
\vspace{-5pt}
\end{figure}
We design synthetic training samples with the goal of further encouraging the model to focus on specific spatial relationships and visual attribute identification. We construct each image as a grid of objects using real-world natural images along with their  segmentation masks from Open-Images-v7~\cite{benenson2022colouring,kuznetsova2020open,openimages} dataset, inspired by prior work using synthetic data to build foundational capabilities in models~\cite{sparkle}. Figure~\ref{fig:syndata} shows examples of synthetic grid visual data and Appendix~\ref{sec:syntheticappendix} show examples of corresponding questions and additional details.

% Our dataset focuses specifically on spatial relationships and visual attribute identification in a controlled setting. Please refer to Appendix~\ref{sec:syntheticappendix} for details.

\section{Results}
\label{sec:results}

This section analyzes the effectiveness of \our. We layer our techniques on top of a LlaVA~\cite{llava} model with Llama 3.1 8B backbone (\texttt{Baseline}) and detail the training setup in Appendix~\ref{sec:trainappendix}. We demonstrate that our approach improves the model's ability to predict visually relevant tokens (Section~\ref{sec:upstream}) and significantly enhances its performance on challenging visual tasks (Section~\ref{sec:spatialevals}).

\subsection{Upstream Analysis}
\label{sec:upstream}
\ignore{What I want to say in this section: i) We use SpatialMM samples for upstream analysis because the text is highly dependent on the visual content, ii) notice a steady decrease in upstream loss with the addition of our technique, iii) visualizing the next token loss values further confirms our intuition that our approach enhances model's ability to generate text related to visual content}

To evaluate the impact of our techniques on a model's ability to integrate visual information during language generation, we conduct an upstream analysis on the SpatialMM dataset~\cite{spatialmm}, that by construction has text with a high dependence on specific visual content, demanding strong visual grounding, making it suitable for this analysis.
\begin{table}
\caption{\label{tab:spatialmm_ntp} Next token prediction (NTP) loss on SpatialMM dataset. We see an improvement as our techniques are introduced. Note that each line is additive, e.g., +\bt\ captures impact of adding \vl\ and \bt\ over the baseline.}
\centering
\vspace{-2pt}
\begin{tabular}{lc}
\toprule
Technique & NTP \\
\midrule
Baseline & 3.59 \\
\addlinespace
+ \vl & 3.54 \\
+ \bt & 3.48 \\
\midrule
+ ind. weights & 3.20 \\
+ AIMv2 encoder (\our) & 2.96 \\
+ \vla & 2.94 \\
\bottomrule
\end{tabular}
\vspace{-10pt}
\end{table}

 Table~\ref{tab:spatialmm_ntp} shows consistent decrease in cross entropy loss for the next token prediction, especially with our proposed enhancements, indicating improved prediction capability.
%
% We observe a clear decrease in this loss metric (Table~\ref{tab:spatialmm_ntp}) as our techniques are introduced, indicating improved next-token prediction accuracy when conditioned on the visual input. 
%
Qualitative visualizations of token-level prediction losses (Appendix~\ref{sec:spatialmmvis}) further support this finding, showing lower values for tokens referencing visual objects, attributes, or relationships within the image.

%\begin{figure}
%\centering
%\includegraphics[width=\columnwidth]{sec/images/spatialmm_ntp.png}
%\caption{\label{fig:spatialmm_viz} This figure visualizes the next-token prediction loss on a per-token basis for our proposed approaches. The visualization highlights improvements in the model's ability to predict tokens corresponding to visual content.}
%\end{figure}

 % This provides strong directional evidence that our approach enhances the model's capability to generate text specifically related to the visual content, moving beyond simple language modeling.

\subsection{Evaluation on SpatialEval Benchmark} % Or \subsubsection
\label{sec:spatialevals}

\ignore{Trends we see in our evaluation of SpatialEval benchmark: i) Consistent improvement as we add our proposed techniques, ii) adding visual loss strengthens model understanding of the content and this amplifies impact of blank tokens where we gain up to 5.5 percentage point improvement, iii) synthetic data further boosts the performance of our training recipe achieving an average improvement of over 9 percentage point. Of particular note here is synthetic data helps with accuracy of \texttt{SpatialReal} subset as well even though synthetic data has a significant domain gap from that task.}
% \begin{table}
% \centering
% \caption{\label{tab:spatialeval_all} Impact of proposed techniques on model performance - we observe consistent improvement in performance as we layer in our innovations. Please note that each line is additive - for instance + \bt\ captures impact of adding \vl\ and \bt\ over the baseline.}
% \begin{tabular}{|c|c|c|c|c|}
% \hline
% Technique & Grid & MazeNav & Map & Real  \\
% \hline\hline
% Baseline & 31.0 & 27.5 & 27.1 & 48.9 \\\hline\hline
% + \vl & 30.4 & 28.4 & 28.6 & 49.8 \\
% + \bt & 31.9 & 30.6 & 32.6 & 50.7\\\hline\hline
% + synthetic &  41.1 & 29.5 & 51.4 & 51.1\\\hline\hline
% + \vla\ & 46.3 & 24.3 & 58.9 & 60.8 \\
% \hline
% \end{tabular}
% \end{table}

We evaluate our proposed techniques on the SpatialEval~\cite{spatialeval} benchmark\footnote{we had to enhance benchmark's prompting and parsing capabilities which we open source as well for reproducibility}, designed to probe spatial understanding capabilities. Our results, summarized in Table~\ref{tab:spatialeval_all}, demonstrate a clear trend of performance improvement across all benchmark subsets with our proposal.

The addition of \vl\ results in a significant improvement in model performance for \texttt{SpatialMap} and \texttt{SpatialReal} subsets of the benchmark. We attribute these improvements to \vl's ability to strengthen the model's understanding of visual content. Adding synthetic data provides consistent improvement across all benchmark subsets, despite minimal overlap with the benchmark tasks. In particular, it helps us narrow the performance gap with the Llama 3.2 11B model~\cite{mllama}, even though our model is smaller (8B vs 11B), utilizes a lower input resolution (e.g., 224x224 vs Llama's higher resolution of 560x560), and employs a simpler modality combination approach (input-layer feature appending vs multi-layer cross-attention). This underscores the synthetic data's effectiveness in enhancing visual grounding by leveraging our proposed training enhancements (Section~\ref{sec:approach}). 

Building on these improvements, we next investigate if our framework can leverage advancements in image encoders to further enhance the LLM's internal visual representation. We find that replacing I-JEPA~\cite{ijepa} with the AM-Radio~\cite{amradio} encoder as the auxiliary image encoder for our visual loss (shown as \vla\ in Table~\ref{tab:spatialeval_all}) yields an additional gain of approximately \fix{9.1 percentage points} on the \texttt{SpatialMap} subset. This enhancement enables our model to match the performance of Llama 3.2 11B on this task, despite our model being smaller, employing a simpler architecture, and using a lower-resolution visual input. 

Our final model, \our, integrates all proposed enhancements -  auxiliary visual loss (\vl), input masking (\bt), the disentangled architecture with independent pathways (Section~\ref{sec:disentangled_weights}), synthetic data (Section~\ref{sec:synthetic}) and an advanced AIM-v2 image encoder~\cite{aimv2}. This configuration yields our strongest results, significantly outperforming the larger Llama 3.2 11B model while being more efficient in terms of model size, architecture, and input resolution. 

\noindent\textbf{Future Work:} An interesting aspect of \our\ design is it's larger capacity and possibility of independently impacting pathways for vision and text. We expect that this departure in model size and capability vs the baseline model could lead to baseline data mixture and hyper-parameter configuration being sub-optimal for \our. For instance, the model's enhanced capacity may make it more susceptible to overfitting on certain peculiarities within the synthetic dataset. This suggests that further tuning of synthetic data generation as well as ablation studies on the data mixture for this model represent a promising avenue for future optimization.

\begin{table}
\centering
\caption{\label{tab:spatialeval_all} Our proposed techniques consistently improve SpatialEvals accuracy across all benchmark subsets. \texttt{Avg.} is calculated as mean of normalized accuracy improvement.}
 \begin{adjustbox}{width=\columnwidth}
\begin{tabular}{lcccccc}
\toprule
Technique & Grid & MazeNav & Map & Real & Avg. \\
\midrule
Baseline & 34.5 & 15.1 & 44.1 & 50.4 & 1 \\
\addlinespace
+ \vl & 31.3 & 17.0 & 50.7 & 56.3 & 1.075 \\
+ \bt & 30.1 & 22.9 & 47.1 & 52.6 & 1.125 \\
+ Synthetic & 41.1 & 26.3 & 55.2 & 53.3 & 1.31 \\
+ \vla\ & 43.5 & 26.7 & 64.1 & 53.3 & 1.385 \\
\midrule
\our & 50.8 & 35.0 & 55.13 & 52.6 & \textbf{1.52} \\
\midrule
\midrule
Llama 3.2 11B~\cite{mllama} & 50.1 & 28.1 & 48.2 & 53.3 & 1.365 \\
\bottomrule
\end{tabular}
\end{adjustbox}
\vspace{-10pt}
\end{table}

\ignore{
\subsection{Future Extensions}
\label{sec:future}

\ignore{We preview possible extensions to our approach based on innovations from the community -- first to reduce the conflict between teaching same set of weights both visual and text representations, we provide each modality independent set of weights while ensuring they can continue to interact with each other through self-attention. second we replace CLIP with a more modern encoder AIMv2. We observe clear improvements from both laying path to rapid advancement in the near future by adopting techniques from the community that can be amplified by our training enhancements.}
We explore an extensions to further improve LLM's internal visual representation. We demonstrate that \vl\ can leverage advancements in image encoders to help the LLM backbone build richer visual representations. We replace I-JEPA~\cite{ijepa} with the AM-Radio~\cite{amradio} encoder as the auxiliary image encoder for our visual loss (shown as \vla\ in Table~\ref{tab:spatialeval_all}) yields an additional gain of approximately \fix{9.1 percentage points} on the \texttt{SpatialMap} subset. This enhancement enables our model to match the performance of Llama 3.2 11B on this task, despite our model being smaller, employing a simpler architecture, and using a lower-resolution visual input. Further investigation of these extensions constitute key directions for our subsequent research.

}
\section{Conclusion}
\label{sec:conclude}

This paper addresses the challenge of weak visual grounding in Multimodal Large Language Models (MLLMs), which often underutilize visual input due to over-reliance on language priors. We provide insights into MLLMs' nascent internal visual representation and propose novel training strategies to strengthen the LLM backbone's visual representation and its ability to leverage this understanding during response generation. The effectiveness of our approach is demonstrated through upstream analysis on the visually rich SpatialMM dataset and accuracy evaluations on the challenging SpatialEvals benchmark. We observe consistent improvement as we layer in our innovations which confirms model's superior visual reasoning capabilities. These results validate our strategy, highlighting the value of simultaneously enhancing core visual understanding and encouraging cross-modal attention as well as paves a scalable path for leveraging advances in image encoders to enhance MLLM's internal visual representation.

\section*{Acknowledgement}
Thanks to Shane Segal for leading the implementation of our disentangled architecture, adapting concepts from Mixture-of-Transformers~\cite{mot} to create specialized processing pathways for the vision and text modalities.
Thanks to Uddeshya Upadhyay for building the pipeline for synthetic data generation and feedback on the early drafts.

{
    \small
    \bibliographystyle{ieeenat_fullname}
    \bibliography{main}

\begin{thebibliography}{24}
\providecommand{\natexlab}[1]{#1}
\providecommand{\url}[1]{\texttt{#1}}
\expandafter\ifx\csname urlstyle\endcsname\relax
  \providecommand{\doi}[1]{doi: #1}\else
  \providecommand{\doi}{doi: \begingroup \urlstyle{rm}\Url}\fi

\bibitem[Agrawal et~al.(2024)Agrawal, Antoniak, Hanna, Bout, Chaplot, Chudnovsky, Costa, Monicault, Garg, Gervet, Ghosh, Héliou, Jacob, Jiang, Khandelwal, Lacroix, Lample, Casas, Lavril, Scao, Lo, Marshall, Martin, Mensch, Muddireddy, Nemychnikova, Pellat, Platen, Raghuraman, Rozière, Sablayrolles, Saulnier, Sauvestre, Shang, Soletskyi, Stewart, Stock, Studnia, Subramanian, Vaze, Wang, and Yang]{pixtral}
Pravesh Agrawal, Szymon Antoniak, Emma~Bou Hanna, Baptiste Bout, Devendra Chaplot, Jessica Chudnovsky, Diogo Costa, Baudouin~De Monicault, Saurabh Garg, Theophile Gervet, Soham Ghosh, Amélie Héliou, Paul Jacob, Albert~Q. Jiang, Kartik Khandelwal, Timothée Lacroix, Guillaume Lample, Diego~Las Casas, Thibaut Lavril, Teven~Le Scao, Andy Lo, William Marshall, Louis Martin, Arthur Mensch, Pavankumar Muddireddy, Valera Nemychnikova, Marie Pellat, Patrick~Von Platen, Nikhil Raghuraman, Baptiste Rozière, Alexandre Sablayrolles, Lucile Saulnier, Romain Sauvestre, Wendy Shang, Roman Soletskyi, Lawrence Stewart, Pierre Stock, Joachim Studnia, Sandeep Subramanian, Sagar Vaze, Thomas Wang, and Sophia Yang.
\newblock Pixtral 12b, 2024.

\bibitem[Assran et~al.(2023)Assran, Duval, Misra, Bojanowski, Vincent, Rabbat, LeCun, and Ballas]{ijepa}
Mahmoud Assran, Quentin Duval, Ishan Misra, Piotr Bojanowski, Pascal Vincent, Michael Rabbat, Yann LeCun, and Nicolas Ballas.
\newblock Self-supervised learning from images with a joint-embedding predictive architecture.
\newblock In \emph{Proceedings of the IEEE/CVF Conference on Computer Vision and Pattern Recognition}, pages 15619--15629, 2023.

\bibitem[Bai et~al.(2024)Bai, Wang, Xiao, He, Han, Zhang, and Shou]{mllmhallucinate}
Zechen Bai, Pichao Wang, Tianjun Xiao, Tong He, Zongbo Han, Zheng Zhang, and Mike~Zheng Shou.
\newblock Hallucination of multimodal large language models: A survey.
\newblock \emph{arXiv preprint arXiv:2404.18930}, 2024.

\bibitem[Benenson and Ferrari(2022)]{benenson2022colouring}
Rodrigo Benenson and Vittorio Ferrari.
\newblock From colouring-in to pointillism: revisiting semantic segmentation supervision.
\newblock \emph{arXiv preprint arXiv:2210.14142}, 2022.

\bibitem[Fini et~al.(2024)Fini, Shukor, Li, Dufter, Klein, Haldimann, Aitharaju, da~Costa, B{\'e}thune, Gan, et~al.]{aimv2}
Enrico Fini, Mustafa Shukor, Xiujun Li, Philipp Dufter, Michal Klein, David Haldimann, Sai Aitharaju, Victor Guilherme~Turrisi da Costa, Louis B{\'e}thune, Zhe Gan, et~al.
\newblock Multimodal autoregressive pre-training of large vision encoders.
\newblock \emph{arXiv preprint arXiv:2411.14402}, 2024.

\bibitem[Grattafiori et~al.(2024)Grattafiori, Dubey, Jauhri, Pandey, Kadian, Al-Dahle, Letman, Mathur, Schelten, Vaughan, et~al.]{mllama}
Aaron Grattafiori, Abhimanyu Dubey, Abhinav Jauhri, Abhinav Pandey, Abhishek Kadian, Ahmad Al-Dahle, Aiesha Letman, Akhil Mathur, Alan Schelten, Alex Vaughan, et~al.
\newblock The llama 3 herd of models.
\newblock \emph{arXiv preprint arXiv:2407.21783}, 2024.

\bibitem[Kuznetsova et~al.(2020{\natexlab{a}})Kuznetsova, Rom, Alldrin, Uijlings, Krasin, Pont-Tuset, Kamali, Popov, Malloci, Kolesnikov, et~al.]{kuznetsova2020open}
Alina Kuznetsova, Hassan Rom, Neil Alldrin, Jasper Uijlings, Ivan Krasin, Jordi Pont-Tuset, Shahab Kamali, Stefan Popov, Matteo Malloci, Alexander Kolesnikov, et~al.
\newblock The open images dataset v4: Unified image classification, object detection, and visual relationship detection at scale.
\newblock \emph{International journal of computer vision}, 2020{\natexlab{a}}.

\bibitem[Kuznetsova et~al.(2020{\natexlab{b}})Kuznetsova, Rom, Alldrin, Uijlings, Krasin, Pont-Tuset, Kamali, Popov, Malloci, Kolesnikov, et~al.]{openimages}
Alina Kuznetsova, Hassan Rom, Neil Alldrin, Jasper Uijlings, Ivan Krasin, Jordi Pont-Tuset, Shahab Kamali, Stefan Popov, Matteo Malloci, Alexander Kolesnikov, et~al.
\newblock The open images dataset v4: Unified image classification, object detection, and visual relationship detection at scale.
\newblock \emph{International journal of computer vision}, 128\penalty0 (7):\penalty0 1956--1981, 2020{\natexlab{b}}.

\bibitem[Liang et~al.(2024)Liang, Yu, Luo, Iyer, Dong, Zhou, Ghosh, Lewis, Yih, Zettlemoyer, et~al.]{mot}
Weixin Liang, Lili Yu, Liang Luo, Srinivasan Iyer, Ning Dong, Chunting Zhou, Gargi Ghosh, Mike Lewis, Wen-tau Yih, Luke Zettlemoyer, et~al.
\newblock Mixture-of-transformers: A sparse and scalable architecture for multi-modal foundation models.
\newblock \emph{arXiv preprint arXiv:2411.04996}, 2024.

\bibitem[Liu et~al.(2023)Liu, Li, Wu, and Lee]{llava}
Haotian Liu, Chunyuan Li, Qingyang Wu, and Yong~Jae Lee.
\newblock Visual instruction tuning.
\newblock \emph{Advances in neural information processing systems}, 36:\penalty0 34892--34916, 2023.

\bibitem[Radford et~al.(2021)Radford, Kim, Hallacy, Ramesh, Goh, Agarwal, Sastry, Askell, Mishkin, Clark, et~al.]{clip}
Alec Radford, Jong~Wook Kim, Chris Hallacy, Aditya Ramesh, Gabriel Goh, Sandhini Agarwal, Girish Sastry, Amanda Askell, Pamela Mishkin, Jack Clark, et~al.
\newblock Learning transferable visual models from natural language supervision.
\newblock In \emph{International conference on machine learning}, pages 8748--8763. PmLR, 2021.

\bibitem[Rahmanzadehgervi et~al.(2024)Rahmanzadehgervi, Bolton, Taesiri, and Nguyen]{vllmblind}
Pooyan Rahmanzadehgervi, Logan Bolton, Mohammad~Reza Taesiri, and Anh~Totti Nguyen.
\newblock Vision language models are blind.
\newblock In \emph{Proceedings of the Asian Conference on Computer Vision (ACCV)}, pages 18--34, 2024.

\bibitem[Ranzinger et~al.(2024)Ranzinger, Heinrich, Kautz, and Molchanov]{amradio}
Mike Ranzinger, Greg Heinrich, Jan Kautz, and Pavlo Molchanov.
\newblock Am-radio: Agglomerative vision foundation model reduce all domains into one.
\newblock In \emph{Proceedings of the IEEE/CVF Conference on Computer Vision and Pattern Recognition}, pages 12490--12500, 2024.

\bibitem[Rasheed et~al.(2024)Rasheed, Maaz, Shaji, Shaker, Khan, Cholakkal, Anwer, Xing, Yang, and Khan]{grand}
Hanoona Rasheed, Muhammad Maaz, Sahal Shaji, Abdelrahman Shaker, Salman Khan, Hisham Cholakkal, Rao~M. Anwer, Eric Xing, Ming-Hsuan Yang, and Fahad~S. Khan.
\newblock Glamm: Pixel grounding large multimodal model.
\newblock In \emph{Proceedings of the IEEE/CVF Conference on Computer Vision and Pattern Recognition (CVPR)}, pages 13009--13018, 2024.

\bibitem[Shi et~al.(2025)Shi, Liu, Wang, Liao, Radhakrishnan, Zhao, Huang, Yin, Sapra, Yacoob, Shi, Catanzaro, Tao, Kautz, Yu, and Liu]{mixtureofencoders}
Min Shi, Fuxiao Liu, Shihao Wang, Shijia Liao, Subhashree Radhakrishnan, Yilin Zhao, De-An Huang, Hongxu Yin, Karan Sapra, Yaser Yacoob, Humphrey Shi, Bryan Catanzaro, Andrew Tao, Jan Kautz, Zhiding Yu, and Guilin Liu.
\newblock Eagle: Exploring the design space for multimodal llms with mixture of encoders, 2025.

\bibitem[Shiri et~al.(2024)Shiri, Guo, Far, Yu, Haf, and Li]{spatialmm}
Fatemeh Shiri, Xiao-Yu Guo, Mona Far, Xin Yu, Reza Haf, and Yuan-Fang Li.
\newblock An empirical analysis on spatial reasoning capabilities of large multimodal models.
\newblock In \emph{Proceedings of the 2024 Conference on Empirical Methods in Natural Language Processing}, pages 21440--21455, 2024.

\bibitem[Tang et~al.(2024)Tang, Qu, Wang, Zhuang, Wu, Ma, Wang, Zheng, Zhao, and Zhao]{sparkle}
Yihong Tang, Ao Qu, Zhaokai Wang, Dingyi Zhuang, Zhaofeng Wu, Wei Ma, Shenhao Wang, Yunhan Zheng, Zhan Zhao, and Jinhua Zhao.
\newblock Sparkle: Mastering basic spatial capabilities in vision language models elicits generalization to composite spatial reasoning.
\newblock \emph{arXiv preprint arXiv:2410.16162}, 2024.

\bibitem[Tong et~al.(2024{\natexlab{a}})Tong, Brown, Wu, Woo, IYER, Akula, Yang, Yang, Middepogu, Wang, et~al.]{cambrian}
Peter Tong, Ellis Brown, Penghao Wu, Sanghyun Woo, Adithya Jairam~Vedagiri IYER, Sai~Charitha Akula, Shusheng Yang, Jihan Yang, Manoj Middepogu, Ziteng Wang, et~al.
\newblock Cambrian-1: A fully open, vision-centric exploration of multimodal llms.
\newblock \emph{Advances in Neural Information Processing Systems}, 37:\penalty0 87310--87356, 2024{\natexlab{a}}.

\bibitem[Tong et~al.(2024{\natexlab{b}})Tong, Liu, Zhai, Ma, LeCun, and Xie]{eyeswideshut}
Shengbang Tong, Zhuang Liu, Yuexiang Zhai, Yi Ma, Yann LeCun, and Saining Xie.
\newblock Eyes wide shut? exploring the visual shortcomings of multimodal llms.
\newblock In \emph{Proceedings of the IEEE/CVF Conference on Computer Vision and Pattern Recognition}, pages 9568--9578, 2024{\natexlab{b}}.

\bibitem[Wang et~al.(2024{\natexlab{a}})Wang, Ming, Shi, Vineet, Wang, Li, and Joshi]{spatialeval}
Jiayu Wang, Yifei Ming, Zhenmei Shi, Vibhav Vineet, Xin Wang, Yixuan Li, and Neel Joshi.
\newblock Is a picture worth a thousand words? delving into spatial reasoning for vision language models.
\newblock In \emph{Advances in Neural Information Processing Systems}, pages 75392--75421. Curran Associates, Inc., 2024{\natexlab{a}}.

\bibitem[Wang et~al.(2024{\natexlab{b}})Wang, Lv, Yu, Hong, Qi, Wang, Ji, Yang, Zhao, XiXuan, et~al.]{cogvlm}
Weihan Wang, Qingsong Lv, Wenmeng Yu, Wenyi Hong, Ji Qi, Yan Wang, Junhui Ji, Zhuoyi Yang, Lei Zhao, Song XiXuan, et~al.
\newblock Cogvlm: Visual expert for pretrained language models.
\newblock \emph{Advances in Neural Information Processing Systems}, 37:\penalty0 121475--121499, 2024{\natexlab{b}}.

\bibitem[Zhang et~al.(2024{\natexlab{a}})Zhang, Yu, Dong, Li, Su, Chu, and Yu]{mllmsurvey}
Duzhen Zhang, Yahan Yu, Jiahua Dong, Chenxing Li, Dan Su, Chenhui Chu, and Dong Yu.
\newblock Mm-llms: Recent advances in multimodal large language models.
\newblock \emph{arXiv preprint arXiv:2401.13601}, 2024{\natexlab{a}}.

\bibitem[Zhang et~al.(2024{\natexlab{b}})Zhang, Li, Li, Ren, Zou, Liu, Huang, Gao, Leizhang, Li, et~al.]{llavaground}
Hao Zhang, Hongyang Li, Feng Li, Tianhe Ren, Xueyan Zou, Shilong Liu, Shijia Huang, Jianfeng Gao, Leizhang, Chunyuan Li, et~al.
\newblock Llava-grounding: Grounded visual chat with large multimodal models.
\newblock In \emph{European Conference on Computer Vision}, pages 19--35. Springer, 2024{\natexlab{b}}.

\bibitem[Zhang et~al.(2024{\natexlab{c}})Zhang, You, Dufter, Zhang, Chen, Chen, Fu, Wang, Chang, Gan, et~al.]{ferret2}
Haotian Zhang, Haoxuan You, Philipp Dufter, Bowen Zhang, Chen Chen, Hong-You Chen, Tsu-Jui Fu, William~Yang Wang, Shih-Fu Chang, Zhe Gan, et~al.
\newblock Ferret-v2: An improved baseline for referring and grounding with large language models.
\newblock \emph{arXiv preprint arXiv:2404.07973}, 2024{\natexlab{c}}.

\end{thebibliography}
}
\clearpage
\setcounter{page}{1}
\maketitlesupplementary

\section{Appendix}
\label{sec:appendix}

\subsection{\vl\ Formulation}
\label{sec:visuallossappendix}

Formally, we introduce an auxiliary vision encoder (say, $\mathcal{A}(\cdot;\theta_A)$) in addition the the standard vision encoder (defined as $\mathcal{G}(\cdot;\theta_G)$ representing CLIP~\cite{clip} in our implementation connected to LLM backbone using MLP connector $\mathcal{M}$). We choose the auxiliary vision encoder to be based on pretrained I-JEPA~\cite{ijepa} model that can extract rich visual representations from the image ($X_\mathcal{I}$). As shown in Figure~\ref{fig:approach}, this auxiliary loss is implemented by predicting the visual features from the LLM backbone ($\mathcal{F}$) and matching them with the auxiliary representation using MSE loss as follows,

\begin{equation}
\begin{aligned}
    &\mathcal{L}_{visualLoss}(\theta_F, \theta_M, \theta_G) = \\
    &\|\mathcal{A}(X_{\mathcal{I}}) - 
    \mathcal{F}\Big(\mathcal{M}\big(\mathcal{G}(X_\mathcal{I})\big) \oplus \mathcal{T}\Big) \odot \texttt{vMask} \|^2
\end{aligned}
\end{equation}

where $\texttt{vMask}$ is the binary mask that only retains the visual features in the output and masks out the textual features. The term $\mathcal{M}\big(\mathcal{G}(X_\mathcal{I})$ represents the visual tokens projected into LLM space and $\oplus \mathcal{T}$ represents the concatenation of textual tokens. The auxiliary loss is combined with the auto-regressive objective for LLM to train the overall MLLM using $\mathcal{L}_{tot}$,

\begin{equation}
\begin{aligned}
    \mathcal{L}_{tot} = \mathcal{L}_{ntp} + \beta \cdot\mathcal{L}_{visualLoss} 
\end{aligned}
\end{equation}

%%
\begin{comment}
\ud{\sout{We retain a standard vision encoder, specifically CLIP~\cite{clip} in our implementation, for processing the input image fed to the MLLM. Crucially, however, we introduce an auxiliary loss term during training, applied directly to the LLM backbone. This loss leverages the output of a separate, semantically powerful visual encoder (I-JEPA~\cite{ijepa} in our work) known for capturing rich visual structure.}}
\end{comment}
%%

\subsection{\bt\ Formulation}
\label{sec:blanktokensappendix}

Formally, we implement \bt\ (\texttt{BlankInputsPartial} from Algorithm~\ref{alg:approach}) as follows: say $T_{in}$ are input text tokens, $b_{id}$ is one of the reserved tokens from LLM vocabulary that we designate as blank token id, and $\mathcal{M}$ is the mask vector determining whether we want to blank an input token, then

\begin{equation}
    T_{inBlank} = 
\begin{cases} 
  T_{in} & \text{if } \mathcal{M} \text{ is True} \\
  b_{Id} & \text{otherwise} 
\end{cases}
\end{equation}

Our specific implementation of this strategy targets the initial tokens during response generation. We consistently blank out the first \texttt{N} text tokens of the input sequence provided to the model during training. 
This deliberate masking at the beginning of the sequence prevents the model from relying on leading textual cues to initiate its response. Instead, the model is forced to formulate its initial coherent thought and begin the generation process based primarily on the processed visual information. This targeted intervention aims to cultivate stronger visual grounding precisely at the critical starting point of generation, encouraging the model's subsequent output to remain more faithful to the visual context. Furthermore, beyond these initial tokens, we randomly blank out a fraction (about 20\%) of the subsequent input tokens to discourage excessive reliance on language context and ensure the model continues to refer back to visual signals throughout its response generation.

\subsection{Synthetic Data Generation}
\label{sec:syntheticappendix}

\begin{figure*}
\centering
\includegraphics[width=0.66\columnwidth]{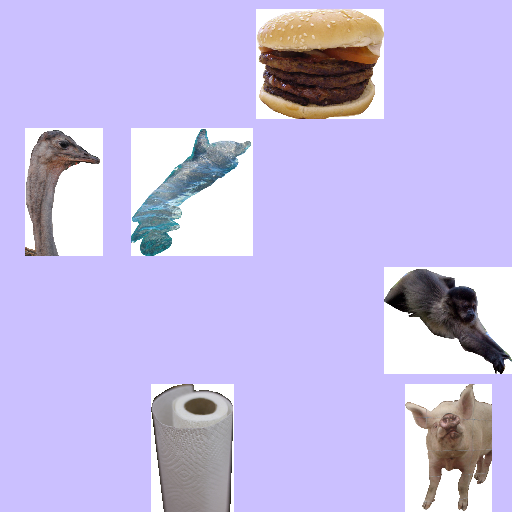}
\includegraphics[width=0.66\columnwidth]{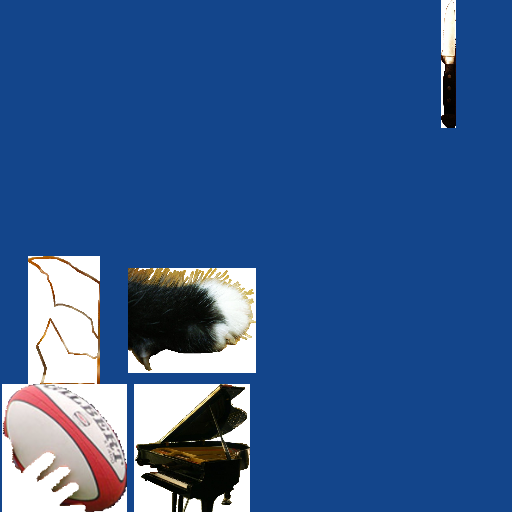}
\includegraphics[width=0.66\columnwidth]{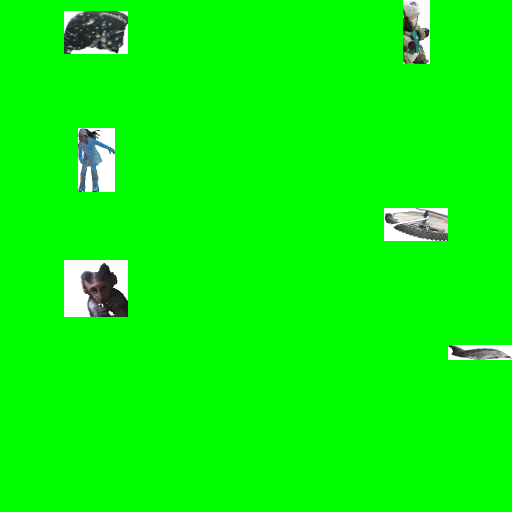}
\caption{\label{fig:syndata_e1} Examples of synthetic visual samples with objects placed on $N \times N$ grid with different backgrounds.}
% \end{figure*}
\includegraphics[width=2\columnwidth]{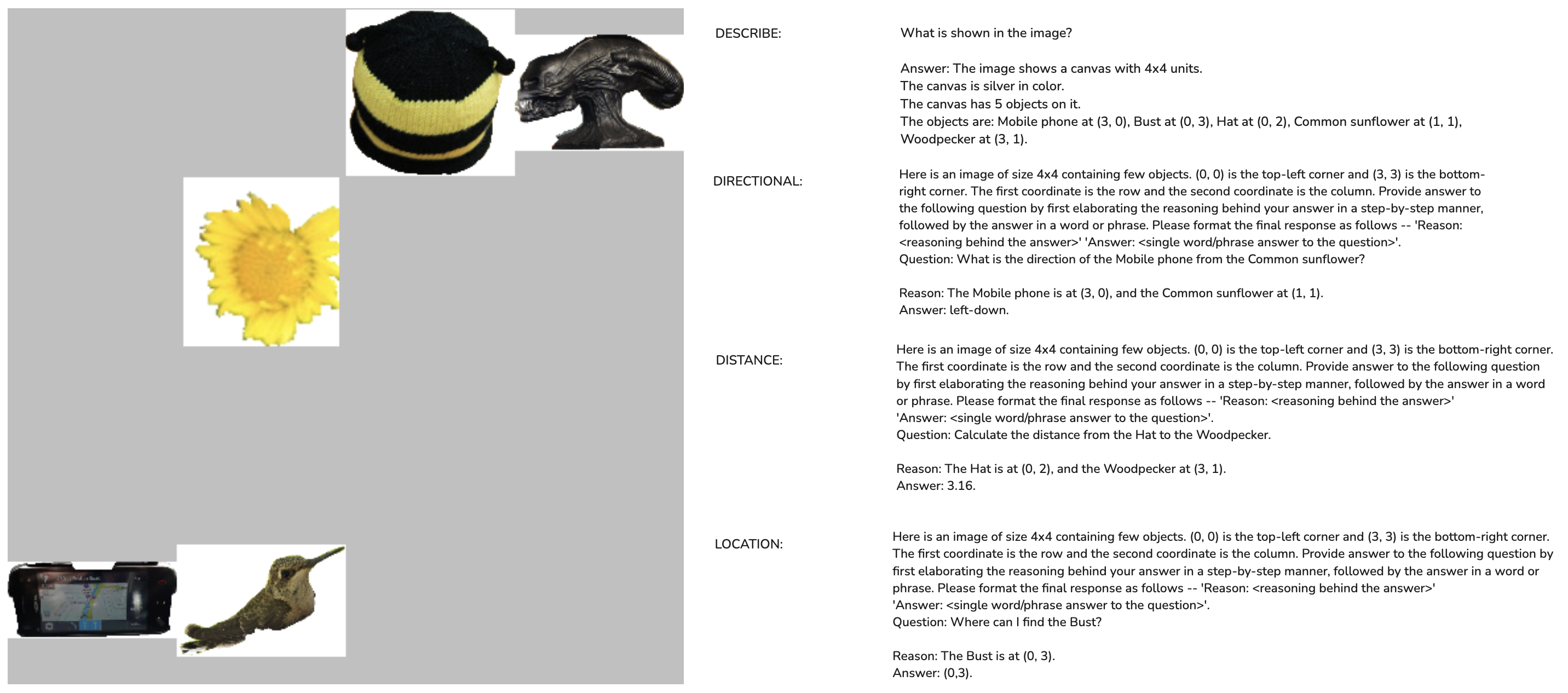}
\caption{\label{fig:syndata_e2} Examples of spatial queries of type \texttt{Describe}, \texttt{Directional}, \texttt{Distance}, \texttt{Location} generated for a visual sample.}
\end{figure*}

The generation process involves programmatically creating visual scenes paired with corresponding questions and answers. We randomly sample object instances from a large public image collection~\cite{openimages}, and place a small number of these objects onto distinct locations within an $N \times N$ grid background. For each generated scene, we automatically formulate basic questions focusing on visual understanding (for example \algcomment{What are the objects in the image?}), relative object direction (for example \algcomment{In which direction is ... ?}) and simple distance (e.g. \algcomment{What is the distance ... ?}). The ground truth answers are derived directly from the programmed spatial layout. 
Figure~\ref{fig:syndata_e1} shows examples of different grid layouts. The grids vary in background colors and grid size granularity, such as 4×4 or 8×8. We formulate 4 types of queries classified as: \texttt{Describe}, \texttt{Directional}, \texttt{Distance}, \texttt{Location}. Figure~\ref{fig:syndata_e2} shows examples of such queries.

Crucially, these generated questions are designed so they can typically only be answered correctly by accurately parsing the visual content of the grid and reasoning about the relative placement of the objects depicted. They inherently resist solutions based purely on language priors or statistical shortcuts. By integrating this synthetic data into our training, we complement our other innovations (\vl, \bt) by providing the model with explicit, targeted tasks that directly exercise and thereby sharpen its visual reasoning and modality alignment capabilities.

\subsection{Training Details}
\label{sec:trainappendix}
\ignore{Explain the following: i) Model Arch is similar to Llava~\cite{llava} with Llama 3.1 8B as the LLM backbone and CLIP as the image encoder, ii) Training setup is similar to Llava where we train the MLP connector only first and then train everything together end-to-end. For visual loss, we use beta of 0.5 and blank out first 5 text tokens, iii) Synthetic data is 25\% of our fine-tuning step.}

\noindent\textbf{Model Architecture.}
Our model architecture is closely based on the LLaVA framework~\cite{llava}. We employ Llama 3.1 8B~\cite{mllama} as the LLM backbone. For visual feature extraction, we utilize the pre-trained CLIP backbone~\cite{clip}.

\noindent\textbf{Training Setup.}
Our training follows a three-stage procedure similar to LLaVA. 
\textit{Stage 1 (Feature Alignment):} We first train only the MLP connector module, keeping both the vision encoder and the LLM backbone frozen. We do not apply \vl\ and \bt\ in this stage.
\textit{Stage 2 (End-to-End Continued Pretraining):} Subsequently, we train the MLP connector, LLM backbone and the image encoder jointly using a mixture of multimodal pretraining dataset. In this stage, we apply \vl\ and \bt\ to encourage LLM to build a rich visual representation.
\textit{Stage 3 (End-to-End Instruction Fine-tuning):} Finally, we fine-tune the MLP connector, LLM backbone and the image encoder to answer questions based on visual and text information.

\noindent\textbf{Technique Implementation \& Data Mixture.}
When incorporating our auxiliary visual loss during training, we set the loss weighting coefficient $\beta = 0.5$. For our input masking technique aimed at reducing language prior reliance, we consistently blank out the first $N=5$ text tokens as well as randomly selected 20\% tokens in the input sequences. During the Stage 3 fine-tuning phase, our data mixture consists of 75\% standard multimodal instruction data and 25\% of our targeted synthetic dataset described in Section~\ref{sec:synthetic}. 

\subsection{Related Work}
\label{sec:relatedworkappendix}

\ignore{Points I want to make: i) rich body of work trying to feed multiple image encoders to LLMs~\cite{cambrian,eyeswideshut,mixtureofencoders} -- marginally helps by providing more information but does not help LLM build better visual representation because the next-token prediction loss is not conducive for integrating the rich visual information into LLM latent space, ii) Adding auxiliary losses for region-level understnding~\cite{grand,ferret2,allseeing,gpt4roi}. Refer to GlaMM~\cite{grand} for detailed analysis of auxiliary visual input. Downside of these approaches is the need for extra visual signals while our work enhances visual representation in a self-supervised manner}

Our approach relates to but diverges from several lines of prior work aimed at improving MLLM visual grounding. We highlight two key areas:

\noindent\textbf{Multiple Image Encoders.} A substantial body of work has explored enhancing visual input by feeding representations from multiple distinct image encoders into the LLM backbone~\cite{cambrian,eyeswideshut,mixtureofencoders}. While providing more diverse visual information can be marginally helpful, this strategy often struggles to significantly improve the LLM's core visual understanding. This limitation arises partly because the standard next-token prediction loss used in LLM training is not inherently conducive to integrating rich, fine-grained visual information deeply into the LLM's latent space.

\noindent\textbf{Auxiliary Visual Losses.} Another relevant direction involves cross-modal attention by incorporating auxiliary losses specifically designed to promote region-level understanding or grounding during training~\cite{grand,ferret2,llavaground}. These methods aim to provide more direct supervision for visual interpretation (GlaMM~\cite{grand} offers a detailed analysis related to auxiliary visual input). However, a common downside of such approaches is the potential need for extra visual signals or annotations (e.g., object locations, regional descriptions) to compute these losses. In contrast, our work focuses on enhancing the LLM's internal visual representation through an auxiliary loss applied in a self-supervised manner, leveraging the structure learned by powerful image encoders without requiring explicit region-level annotations for the auxiliary task itself.

\subsection{Next-token Prediction Visualization}
\label{sec:spatialmmvis}
\begin{figure*}
\centering
\captionsetup{width=0.9\textwidth}
\begin{measuredfigure}
\includegraphics[width=0.9\textwidth]{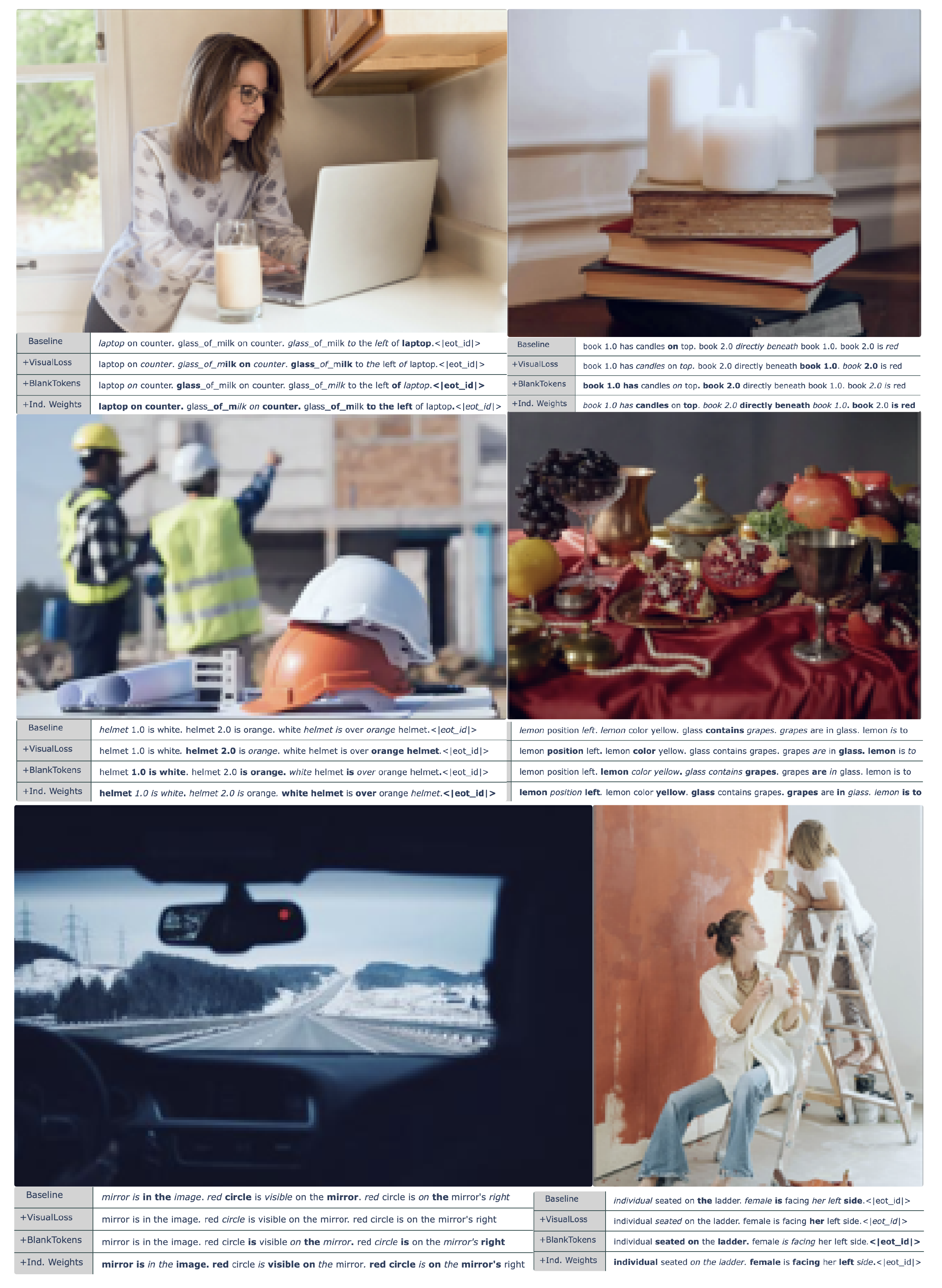}
\caption{\label{fig:spatialmmvis} This figure visualizes the next-token prediction loss on a per-token basis for our proposed approaches. The visualization highlights improvements in the model's ability to predict tokens corresponding to visual content. In the figure, \textbf{bold} text represents the lowest cross entropy loss when comparing across the model variants.}
\end{measuredfigure}
\end{figure*}

We make the following observations regarding the next-token prediction accuracy shown in the figure:
\begin{itemize}
    \item \textbf{Higher-quality captions:} We observe that the first few tokens are consistently more accurate for our model compared to the baseline. This demonstrates that our \bt\ intervention enhances the model's ability to compose text grounded in the visual content from the outset.
    \item \textbf{Higher confidence prediction of visually relevant text tokens:} We observe that tokens pertaining to visual concepts (such as objects, people, orientation, shape, and color) are predicted with higher confidence by the model trained with our innovations.
    \item \textbf{Focus on enriching cross-modal alignment:} The baseline model achieves its highest prediction accuracy primarily for language-centric tokens like \texttt{contains}, \texttt{in the}, and \texttt{side}. This further supports the conclusion that our approach's primary impact is on improving the generation of visually grounded text requiring reference to visual input and reasoning.
\end{itemize}


\end{document}